\documentclass[pdflatex,sn-nature]{sn-jnl}

\usepackage{graphicx}%
\usepackage{multirow}%
\usepackage{amsmath,amssymb,amsfonts}%
\usepackage{amsthm}%
\usepackage{mathrsfs}%
\usepackage[title]{appendix}%
\usepackage{xcolor}%
\usepackage{textcomp}%
\usepackage{manyfoot}%
\usepackage{booktabs}%
\usepackage{algorithm}%
\usepackage{algorithmicx}%
\usepackage{algpseudocode}%
\usepackage{listings}%
\usepackage{geometry} 

\theoremstyle{thmstyleone}%

\theoremstyle{thmstyletwo}%

\theoremstyle{thmstylethree}%

\begin{document}

\title[Article Title]{Universal and efficient graph neural networks with dynamic attention for machine learning interatomic potentials}

\author[1]{\fnm{Shuyu} \sur{Bi}}
\author[1]{\fnm{Zhede} \sur{Zhao}}
\author[1]{\fnm{Qiangchao} \sur{Sun}}

\author*[1]{\fnm{Tao} \sur{Hu}}\email{taohu@shu.edu.cn}
\author[1]{\fnm{Xionggang} \sur{Lu}}
\author*[1]{\fnm{Hongwei} \sur{Cheng}}\email{hwcheng@shu.edu.cn}

\affil[1]{\orgdiv{School of Materials Science and Engineering \& State Key Laboratory of Advanced
Refractories \& State Key Laboratory of Advanced Special Steels}, \orgname{Shanghai University}, \orgaddress{\city{Shanghai}, \postcode{200444}, \country{P. R. China}}}

\abstract{The core of molecular dynamics simulation fundamentally lies in the interatomic potential. Traditional empirical potentials lack accuracy, while first-principles methods are computationally prohibitive. Machine learning interatomic potentials (MLIPs) promise near-quantum accuracy at linear cost, but existing models still face challenges in efficiency and stability. We presents Machine Learning Advances Neural Network (MLANet), an efficient and robust graph neural network framework. MLANet introduces a dual-path dynamic attention mechanism for geometry-aware message passing and a multi-perspective pooling strategy to construct comprehensive system representations. This design enables highly accurate modeling of atomic environments while achieving exceptional computational efficiency, making high-fidelity simulations more accessible. Tested across a wide range of datasets spanning diverse systems, including organic molecules (e.g., QM7, MD17), periodic inorganic materials (e.g., Li-containing crystals), two-dimensional materials (e.g., bilayer graphene, black phosphorus), surface catalytic reactions (e.g., formate decomposition), and charged systems, MLANet maintains competitive prediction accuracy while its computational cost is markedly lower than mainstream equivariant models, and it enables stable long-time molecular dynamics simulations. MLANet provides an efficient and practical tool for large-scale, high-accuracy atomic simulations.}

\maketitle

\section*{Introduction}\label{sec1}

Molecular dynamics (MD) simulations have emerged as a cornerstone methodology for investigating dynamical processes across a wide spectrum of scientific disciplines, including chemistry, biology, physics, and materials science, effectively bridging atomic-scale interactions to macroscopic functional properties\cite{MD1}. At the heart of MD resides the calculation of interatomic potentials, which has traditionally relied on empirical potential functions\cite{MD2, MD3, MD4, MD5, MD6, MD7, MD8, MD9}. While these classical force fields offer O(N) computational efficiency, their accuracy is fundamentally limited by predefined functional forms and manual parameterization, often failing to capture complex many-body interactions\cite{unke_machine_2021,mueller_machine_2020,shapeev_moment_2016,Stocker_2022} .
To achieve quantum-mechanical accuracy, ab initio molecular dynamics (AIMD) methods based on density functional theory (DFT) have been extensively utilized and applied \cite{kohn_self-consistent_nodate}. Although DFT provides rigorous first-principles precision, its prohibitive O(N³) computational scaling restricts simulations to systems below 1,000 atoms and picosecond timescales, making large-scale phenomena like defect dynamics or catalytic reactions computationally intractable \cite{Plimpton_1993, niethammer_ls1_2014}. For such cases, parallelized domain decomposition and specialized algorithms can extend the accessible scale of classical MD , albeit without quantum accuracy.

The advent of machine learning interatomic potentials (MLIPs) has revolutionized this landscape, offering near-DFT accuracy with linear computational scaling. By training on ab initio datasets, MLIPs infer energies and forces through data-driven representations, bypassing explicit electronic structure calculations\cite{reiser_graph_2022, jmi.2025.17} . 
Early invariant MLIPs like CGCNN\cite{xie_crystal_2018}, SchNet\cite{schutt_schnet_2018} and DimeNet\cite{gasteiger_directional_2022} relied on handcrafted geometric features (e.g., interatomic distances, angles), but their fixed descriptors struggled to distinguish configurations with identical local geometries\cite{unke_physnet_2019,lu_pre-training_2024} . Subsequent advances in MLIPs, such as ALIGNN\cite{choudhary_atomistic_2021} and M3GNet\cite{chen_universal_2022}, incorporated higher-body terms like dihedral angles, improving angular resolution while maintaining E(3) invariance .
A paradigm shift occurred with the introduction of SE(3)-equivariant architectures, which explicitly preserve rotational and translational symmetries through tensor representations . Models like Allegro\cite{musaelian_learning_2023}, NequIP\cite{batzner_e3-equivariant_2022} and MACE\cite{Batatia2022mace} leverage spherical harmonics and irreducible representations (irreps) to encode atomic environments, achieving sub-meV/atom errors on benchmarks through tensor product operations. These equivariant MLIPs demonstrate superior data efficiency, requiring orders of magnitude fewer training samples than invariant counterparts. Recent extensions, such as eSCN, reduce the computational overhead of high-order irreps (Lmax) from $O(L^6)$ to $O(L^3)$, enabling maximum rotation order up to $l_{max}=8$ while retaining quantum accuracy\cite{fu2025learningsmoothexpressiveinteratomic} .

Despite these advances, critical challenges persist: Computational complexity remains a significant barrier, as high-order tensor products in equivariant models incur substantial costs, limiting practical $l_{max}$ values and forcing trade-offs between angular resolution and speed \cite{Plimpton_1993, niethammer_ls1_2014}. This computational burden mirrors the traditional trade-offs seen in classical force field development, where higher accuracy typically came at the expense of performance \cite{slim_toward_2008,liu_large-scale_2023}. Parallelization bottlenecks further compound these issues - the message-passing mechanism in MLIPs expands the effective communication radius to r$_c$ × N$_{layers}$ in spatial decomposition schemes, causing redundant computations and poor scaling beyond 32,000 atoms even on 64 GPUs\cite{batatia2025foundationmodelatomisticmaterials}.

In this work, we propose the MLANet framework. The model employs a geometry-aware dual-path dynamic attention mechanism within its equivariant message-passing layers, which adaptively modulates interatomic interactions based on both geometric and chemical features, alongside a physics-informed multi-perspective pooling strategy that mitigates information loss by combining complementary pooling operations. These designs are engineered to balance high accuracy with superior computational efficiency, addressing a central challenge in modern equivariant graph neural networks. The following sections systematically demonstrate its performance: validating its accuracy and stability in predicting energies and forces on organic molecular datasets such as QM7\cite{blum, rupp} and MD17\cite{MD17}; examining its fitting and generalization capabilities for energy-volume relationships, atomic pair interactions, and long-range van der Waals forces in inorganic material systems including SiO$_2$\cite{erhard_machine-learned_2022}, Ge-Sb-Te\cite{zhou_device-scale_2023}, and black phosphorus\cite{deringer_general-purpose_2020}; and assessing its application potential in systems such as two-dimensional materials (e.g., bilayer graphene\cite{ying_effect_2024}) and surface catalysis (e.g., formic acid decomposition\cite{batzner_e3-equivariant_2022}). In addition, the model demonstrates competitive performance in charged systems and quantum chemical property predictions for organic molecules\cite{ramakrishnan2014quantum, zou_deep_2023, ko_fourth-generation_2021}. Our results show that while maintaining accuracy comparable to mainstream equivariant models, MLANet achieves a significant improvement in computational efficiency, providing an efficient and reliable solution for high-accuracy atomic simulations across scales.

\begin{figure}[htbp]
\centering
\includegraphics[width=0.9\textwidth]{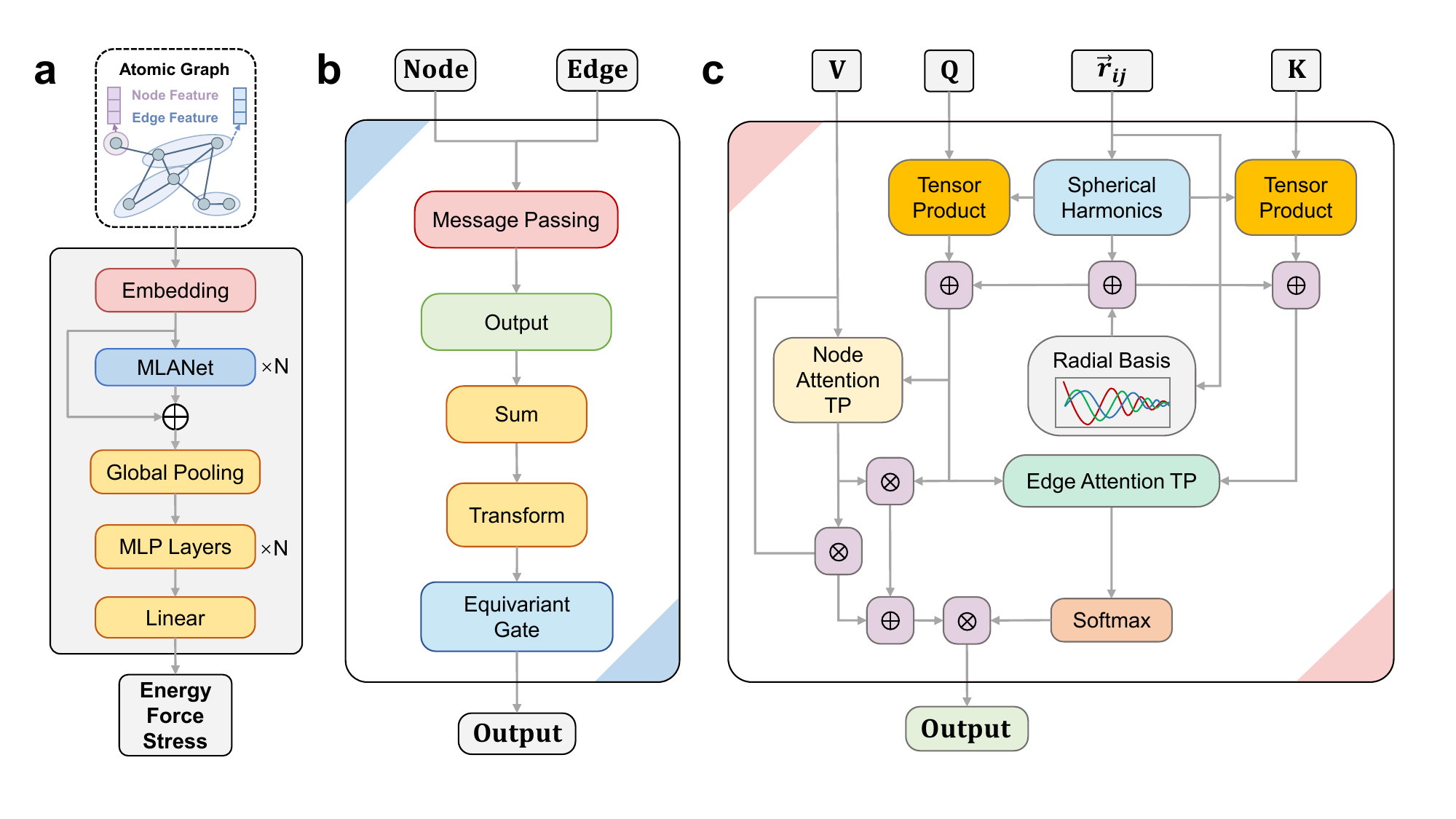}
\caption{\textbf{The architecture of MLANet.} \textbf{a} Overall Model Architecture. \textbf{b} Internal Structure of MLANet.\textbf{c} Message Passing Layer of MLANet.}
\label{fig1}
\end{figure}

\section*{Results}
\subsection*{Model Architecture}
The core of our framework builds upon an equivariant graph neural network with two specialized components (Fig. \ref{fig1}): 
Equivariant Message Passing with geometry-aware dual-path dynamic Graph Attention that combines tensor products with spherical harmonics to maintain $SE(3)$-equivariance during neighbor aggregation and adaptively modulate message flows based on both geometric distances and chemical embeddings;
multi-perspective Pooling that concatenate additive, mean, and max graph-level features before final prediction.

\subsubsection*{Graph Construction}
    The atomic system is represented as a graph $G = (V, E, C)$, where:
$V = \{v_i\}_{i=1}^N$ is the node feature vector formed by concatenating atomic type embeddings:
\begin{equation}
        v_i = \operatorname{Embedding}(z_i)
        \end{equation}
        where $z_i$ denotes the atomic type.$E = \{e_{i,j}\}$ represents edge features, and the Bessel function basis is defined as:
       \begin{equation}
        \operatorname{RBF}_n(x) = \sqrt{\frac{2}{r_{\text{cut}}}} \frac{j_n\left(\frac{\alpha_{n,m}x}{r_{\text{cut}}}\right)}{|j_n(\alpha_{n,m})|}
        \end{equation}
        where $j_n$ is the $n$-th order Bessel function.
            $\alpha_{n,m}$ is the $m$-th root of $j_n$.
            $r_{\text{cut}}$ is the cutoff radius.
            The normalization ensures orthonormality within the cutoff radius.

        The edge construction method accounts for periodic boundary conditions, where $C \in \mathbb{R}^{3\times3}$ is the optional unit cell matrix. We do not explicitly provide the unit cell as model input; it is implicitly considered through the edge construction process.

\subsubsection*{Equivariant Message Passing and Update}

The message passing layer in MLANet implements a geometry-aware dual-path dynamic attention mechanism. Its operation proceeds through the following sequence of mathematical operations:

    The spherical harmonics up to maximum rotation order $l_{max}$ are computed for the normalized direction vectors $\hat{r}_{ij}$ between neighboring atoms:
    \begin{equation}
        Y_1(\hat{r}_{ij}) = \operatorname{SH}(\hat{r}_{ij})
    \end{equation}

    Edge features are transformed by processing raw edge attributes through head-specific linear layer $W_{\text{edge\_raw}}$,
        processing spherical harmonics through separate linear layer $W_{\text{edge\_sh}}$ and
        combining both components additively:
    \begin{equation}
        e_{ij} = W_{\text{edge\_raw}}(\text{edge\_attr}) + W_{\text{edge\_sh}}(Y_1(\hat{r}_{ij}))
    \end{equation}

    Node features are linearly transformed into queries ($q_i$) and keys ($k_j$), with edge features added to both:
    \begin{equation}
        q_i = W_q x_i + e_{ij}, \quad k_j = W_k x_j + e_{ij}
    \end{equation}

    The attention weights are computed by taking the tensor product (TP) between queries and keys,
        scaling by temperature parameter $\tau$ and
        applying softmax normalization over neighboring atoms $j$:
    
    \begin{equation}
        \alpha_{ij} = \operatorname{softmax}\left(\frac{\text{TP}(q_i, k_j)}{\tau}\right)
    \end{equation}
    
A gating mechanism is computed by transforming neighbor features with $v_j = W_v x_j$,
        computing tensor product with $q_i$ and
        applying sigmoid activation $\sigma$:

    \begin{equation}
        \beta_{ij} = \sigma(\operatorname{TP}(q_i, v_j))
    \end{equation}

    The final message combines transformed neighbor features ($v_j$) weighted by gate $\beta_{ij}$ and
        query features ($q_i$) weighted by complementary gate $(1-\beta_{ij})$, and Weighted by their respective attention scores $\alpha_{ij}^h$:
    \begin{equation}
        m_{ij} = \beta_{ij} \cdot v_j + (1-\beta_{ij}) \cdot q_i
    \end{equation}

    \begin{equation}
        m_i =  \alpha_{ij} \cdot m_{ij}
    \end{equation}

    The final node update consists of linear transformation of aggregated messages ($W_{\text{trans}}$),
        gating operation (described below) and
        residual connection with input features:
    
    \begin{equation}
        x_i^{(l+1)} = \operatorname{Gate}\left(W_{\text{trans}} m_i\right) + x_i^{(l)}
    \end{equation}
    
    The gating mechanism plays a crucial role in the node feature update process, ensuring that the model can effectively apply non-linear transformations while preserving $SE(3)$-equivariance. In our architecture, we employ SiLU (Swish) activation for scalar features and non-scalar features, which enhances model expressiveness while maintaining stable gradient flow.

    The mathematical formulation of the gate operation is defined as:

    \begin{equation}
        \operatorname{Gate}(x) =  (\phi_{\text{scalars}}(x_{\text{s}})) \oplus \left( (\phi_{\text{gates}}(x_{\text{n}})) \odot x_{\text{non-scalars}}\right)
    \end{equation}
    where $x_{\text{s}}$ denotes the scalar features (irreps with $l=0$), processed through $\text{Linear}_{\text{s}}$ with SILU activation.
        $x_{\text{n}}$ represents higher-order features (irreps with $l>0$).
        $\phi $ denotes SILU activation.

        $\odot$ denotes element-wise multiplication.
        $\oplus$ indicates concatenation of processed components.
    The activation functions are defined as:
    \begin{equation}
    \operatorname{SiLU}(x) = x \cdot \sigma(x) \\
    \end{equation}
    where $\sigma(x)$ is the sigmoid function.

\subsubsection*{Physics-Informed Multi-Perspective Pooling}
For each graph $G$ in batch, we compute three pooling operations and concatenate them:

\begin{equation}
\begin{aligned}
&v_{\text{add}} = \sum_{i\in G} x_i, \quad v_{\text{mean}} = \frac{1}{|G|}\sum_{i\in G} x_i \\
&v_{\text{max}} = \max_{i\in G} x_i, \quad v_{\text{pooled}} = v_{\text{add}} \oplus v_{\text{mean}} \oplus v_{\text{max}}
\end{aligned}
\end{equation}

Additive Pooling preserves extensive properties (e.g., total energy as a sum), Mean Pooling captures intensive, size-invariant properties and Max Pooling highlights critical local environments (e.g., active sites).
By concatenating these representations and learning their optimal combination, MLANet constructs a richer, more robust global descriptor. This mitigates information loss, enhances generalization, and eliminates the need for complex, computationally heavy post-processing layers to achieve accurate energy predictions.

\subsubsection*{Prediction Heads}
The architecture uses separate prediction heads for Energy ($E\in\mathbb{R}$), Forces ($\vec{f}_i\in\mathbb{R}^3$) and Stress ($\sigma\in\mathbb{R}^6$):

    \begin{equation}
    E = \operatorname{MLP}_{\text{energy}}(v_{\text{pooled}})
    \end{equation}

    \begin{equation}
    \vec{f}_i = \operatorname{MLP}_{\text{forces}}(x_i \oplus v_{\text{pooled}}[b_i])
    \end{equation}

    \begin{equation}
    \sigma = \operatorname{MLP}_{\text{stress}}(v_{\text{pooled}})
    \end{equation}

\begin{figure}[htpb!]
\centering
\includegraphics[width=0.9\textwidth]{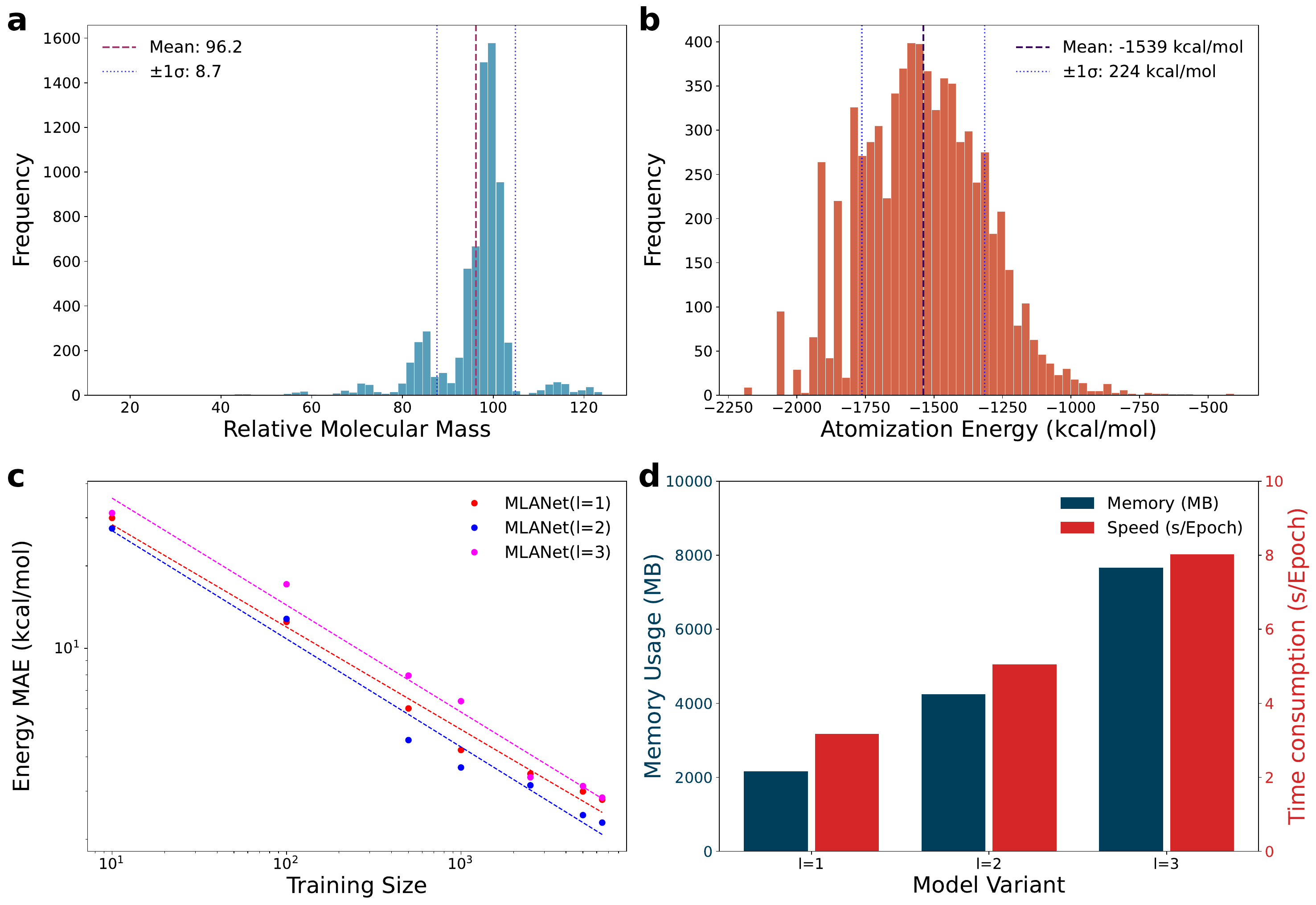}
\caption{\textbf{Learning Curves and Model Efficiency of MLANet on QM7.} \textbf{a} Histogram of molecular masses in the QM7 dataset, \textbf{b} histogram of atomization energies. \textbf{c} Log–log learning curve of energy prediction MAE versus training set size, \textbf{d} memory usage and training time per epoch for MLANet with maximum rotation order $l_{max} \in \{1, 2, 3\}$.}
\label{fig2}
\end{figure}

\subsection*{Learning curves and scaling behavior of MLANet}

To investigate the influence of the angular momentum parameter and the training set size on the model's learning capacity and efficiency, we conduct training experiments using different angular momentum parameters and varying training set sizes. 
These experiments are performed on two distinct datasets: the QM7 dataset\cite{blum, rupp}, representing non-periodic small organic molecular systems, and the Li-containing subset of the Mptrj dataset\cite{deng_2023_chgnet}, representing inorganic periodic crystals.

We first evaluate MLANet on the QM7 organic small molecule dataset. Multiple models are trained using varying numbers of training samples (10, 100, 500, 1000, 2500, 5000, and 6449) and test on a fixed set of 716 samples, with the learning target being the atomization energy. 
The distribution of relative molecular mass (Fig. \ref{fig2}a) and atomization energy (Fig. \ref{fig2}b) within the QM7 dataset is illustrated. The similarity in the distribution of these small molecules suggests that the dataset is relatively homogeneous, which facilitates model learning. 
The log-log learning curves are presented (Fig. \ref{fig2}c), plotting the mean absolute error (MAE) of the predicted atomization energy against the training set size for MLANet models with maximum rotation order $l_{max} \in \{1, 2, 3\}$. The MAE for all three models shows a clear linear decreasing trend on the log-log scale, consistent with typical power-law scaling. Notably, the model with $l=2$ achieves a lower MAE than the model with $l=3$ across most data regimes, indicating that a higher angular momentum cutoff does not guarantee better performance for energy prediction, especially with limited training data.
Finally, the computational cost is reported (Fig. \ref{fig2}d), showing the memory usage and training time per epoch for different $l$ values. As expected, both memory consumption and training time increase substantially with higher $l$, highlighting the trade-off between model expressivity and computational efficiency.

\begin{figure}[b]
\centering
\includegraphics[width=0.9\textwidth]{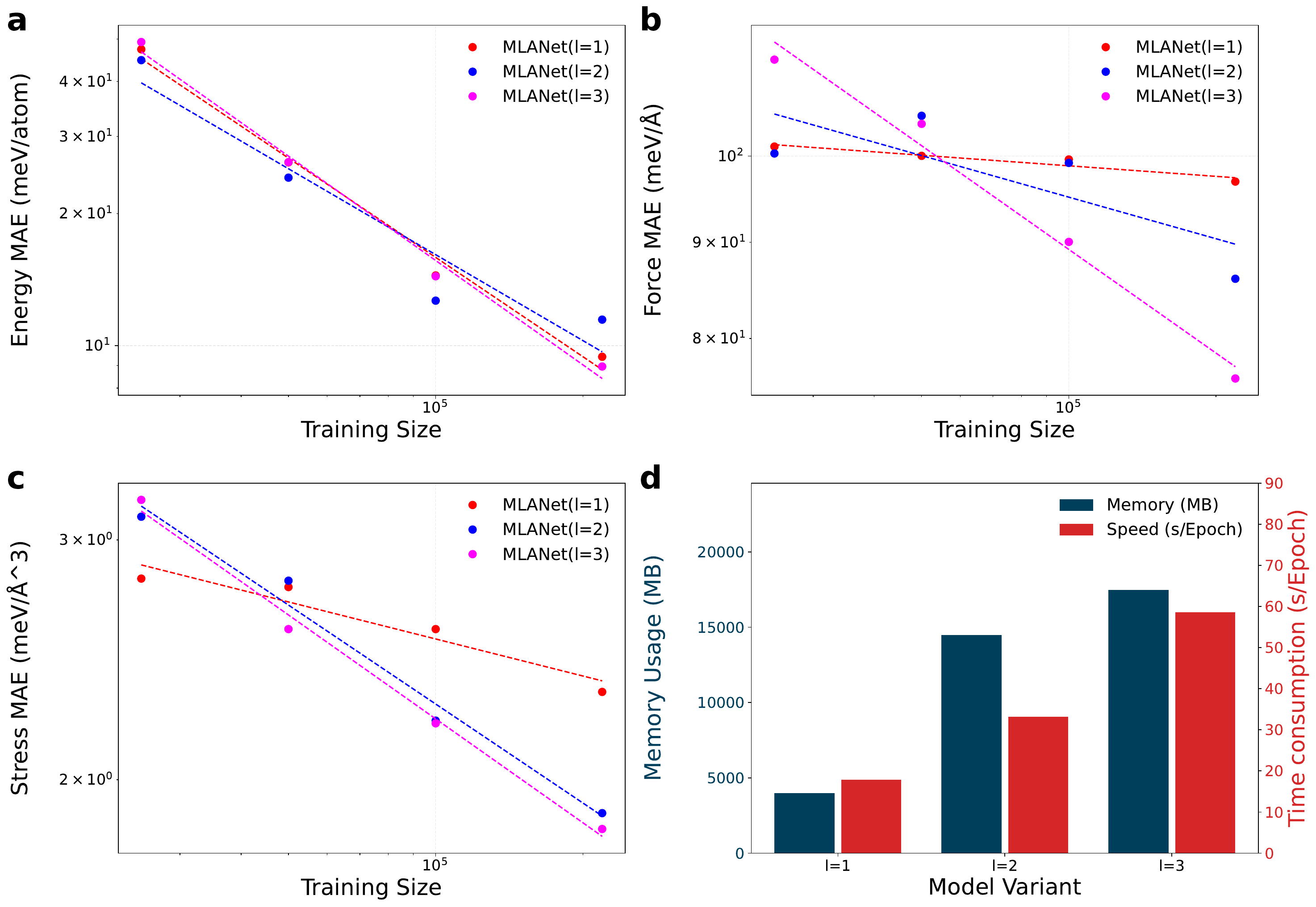}
\caption{\textbf{Learning curves and computational efficiency for MLANet on the lithium-containing subset of Mptrj.} \textbf{a,\ b,\ c} Log-log plots show the scaling of energy, force, and stress prediction errors (MAE) with training set size for MLANet variants with maximum rotation order $l_{max} \in \{1, 2, 3\}$. \textbf{d} The  panel compares memory usage and training speed per epoch across model variants.
}
\label{fig3}
\end{figure}

We then evaluate MLANet on the lithium-containing subset of the Mptrj dataset. Models are trained for 500 epochs using varying numbers of training samples (25,000, 50,000, 100,000, and 219,033) and tested on a fixed set of 1,449 samples. The learning targets are the energy, forces, and stress (Fig. S1 in the Supplementary Information(SI)).
The learning curves and computational efficiency of MLANet on this subset are presented (Fig. \ref{fig3}a to \ref{fig3}c). The panels show the scaling of the prediction mean absolute error (MAE) for energy, forces, and stress with respect to the training set size for MLANet variants with maximum rotation order $l_{max} \in \{1, 2, 3\}$. All three properties exhibit the same trend: the performance of MLANet becomes more sensitive to the dataset size as $l$ increases. With larger datasets, higher $l$ variants demonstrate stronger fitting capability and achieve lower MAEs. Conversely, with smaller datasets, they are more prone to overfitting, resulting in higher MAEs. This indicates that flexibly selecting $l$ based on the available data scale and the specific system is crucial for achieving an optimal balance between efficiency and accuracy. The final panel compares the memory usage and training speed per epoch for each model variant (Fig. \ref{fig3}d), which follows the expected pattern where both computational cost and memory consumption increase with $l$.

\subsection*{Inorganic systems}

\begin{figure}[b]
\centering
\includegraphics[width=0.9\textwidth]{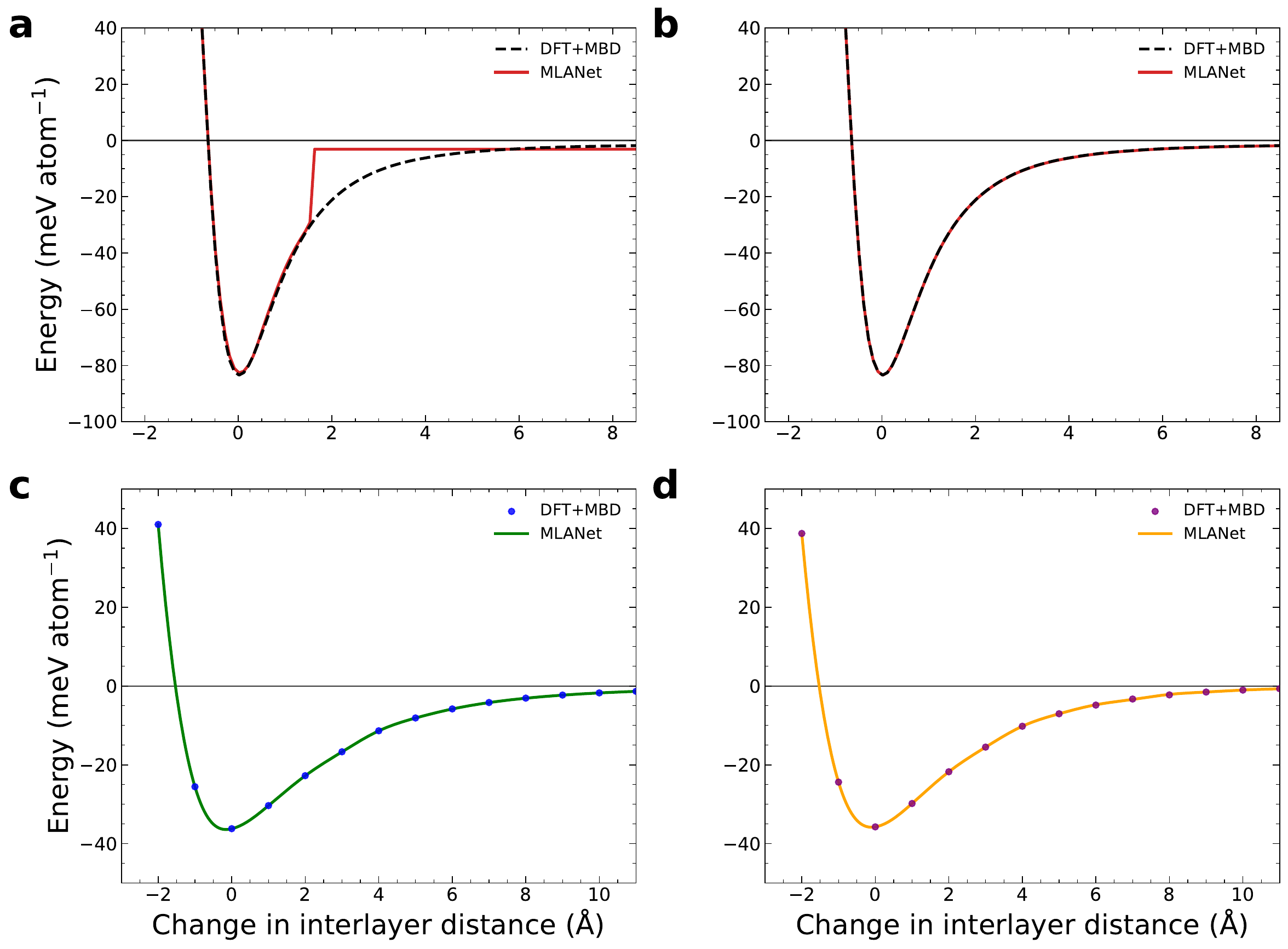}
\caption{\textbf{Exfoliation of phosphorene. }
\textbf{a,\ b} Exfoliation curves of black phosphorene predicted by the MLANet model without (\textbf{a}) and with (\textbf{b}) the long‑range interaction term (red lines), 
compared with DFT + MBD reference data (black dashed lines). \textbf{c,\ d} Exfoliation curves of Hittorf’s phosphorus (P2/c and P2/n structures) predicted by the MLANet model including the long‑range interaction term compared with DFT + MBD reference data points The plots show the energy evolution with interlayer distance.
}
\label{fig5}
\end{figure}

To demonstrate the fitting capability of the model, we train it on several inorganic material systems.
The energy-volume curves for several silica polymorphs calculated using DFT\cite{erhard_machine-learned_2022} (with the SCAN functional) and MLANet are presented (Fig. S2 in the SI). Accurate reproduction of these curves is a key initial test for evaluating a potential's thermodynamic behavior, as the 0 K energy-volume diagram is critical for predicting phase transitions. MLANet (colored points) shows excellent agreement with the SCAN-DFT reference (black curves), confirming its high-precision predictive performance.
The potential energy scans for atomic pair interactions calculated using DFT\cite{zhou_device-scale_2023} and MLANet are presented (Fig. S3 in the SI). The results cover scans of isolated dimers for all atomic pair types in the Ge–Sb–Te system. MLANet (red curves) reproduces the DFT calculations (white circles) accurately, showcasing MLANet's high-precision predictive capability for intermetallic interactions.
The binding energy--distance curves for bilayer black phosphorus exfoliation calculated using DFT+MBD\cite{deringer_general-purpose_2020} and MLANet are presented in Fig. \ref{fig5}a and \ref{fig5}b. Testing on this dataset reflects the model's comprehensive predictive ability for both short-range and long-range interactions. By incorporating a long-range interaction term, MLANet (red curve) precisely reproduces the DFT+MBD reference data (black dashed curve), demonstrating the long-range awareness of the model. Furthermore, as shown in Fig. \ref{fig5}c and \ref{fig5}d, MLANet with the long-range term also accurately captures the exfoliation energetics of two Hittorf's phosphorus polymorphs (P2/c and P2/n structures), further validating its transferability across different phosphorene systems.

\subsection*{Small organic molecules}
To evaluate the model's predictive accuracy, inference speed, and stability, we conducte further tests on a series of aperiodic organic small-molecule datasets.

\begin{table*}[h]
		\small
		\centering
    \setlength{\tabcolsep}{12pt} 
		\caption{\textbf{Results on the QM7 datasets.} Boldface indicates the best performance. Errors shown in parentheses are obtained as the standard deviations from ten-fold cross validation.}
			\begin{tabular}{ll}
				\toprule
				\textbf{Method} &	\textbf{MAE} [kcal/mol] \\
				\midrule
				AttentiveFP\cite{xiong_pushing_2020}  &  			66.2  (2.8)     \\
				DMPNN \cite{yang_analyzing_2019} &  			105.8  (13.2)     \\	
				DTNN \cite{wu_moleculenet_2018} &  	 		 8.2  (3.9)    \\
				GraphConv \cite{Kipf2016SemiSupervisedCW}   & 118.9  (20.2)   	\\
				
				GROVER (base)\cite{0b32fdade44242898a13df078619c777} &  			72.5 (5.9)      \\	
				MPNN \cite{pmlr-v70-gilmer17a} &  			113.0  (17.2)      \\				
				
				N-GRAM\cite{NEURIPS2019_2f3926f0} &  			125.6  (1.5)      \\		
				PAGTN (global)\cite{chen2019pathaugmentedgraphtransformernetwork}  &  			47.8 (3.0)      \\	
				
				PhysChem\cite{NEURIPS2021_884d247c}  &  			59.6 (2.3)      \\

				SchNet\cite{schutt_schnet_2018}  &  			74.2  (6.0)      \\
				
				Weave \cite{kearnes_molecular_2016} &  			59.6 (2.3)      \\
				
				GGST+EK\cite{10.5555/3600270.3602461}&  		11.3  (0.6)    \\	
				
				$2$GGST+EK\cite{10.5555/3600270.3602461} &  	 		3.4  (0.3)     \\	
        MLANet &  	 		\textbf{3.07}  (\textbf{0.09})     \\

				\bottomrule
			\end{tabular}
		
		\label{qm7}
	\end{table*}

Building upon the learning curve discussed above, we present a comparison of the mean absolute error (MAE) and standard deviation for atomization energy predictions across different models on the QM7 dataset (Table. \ref{qm7}). MLANet achieves the lowest error and smallest standard deviation, demonstrating its superior accuracy in predicting atomization energies.

In addition to the classical tasks of energy and force prediction for machine-learned interatomic potentials, we also evaluate MLANet on the QM9\cite{ramakrishnan2014quantum} and QM9S\cite{zou_deep_2023} dataset for predicting molecular quantum chemical properties (Table. S1 in the SI and Table. \ref{tab:QM9S}). MLANet achieves accuracy comparable to models evaluated on the QM9 dataset, as well as to other models except those specifically designed for predicting the QM9S dataset (DetaNet and EnviroDetaNet), demonstrating its feasibility in this domain.

\begin{table*}[h]
  \centering
  \caption{\textbf{Model performance (MAE) on the QM9S dataset.}}
  \label{tab:QM9S}
  \resizebox{\textwidth}{!}{
  \begin{tabular}{llllllll}
    \toprule
           
           & SchNet\cite{schutt_schnet_2018}
           & DimeNet++\cite{gasteiger_dimenetpp_2020}
           & PaiNN\cite{pmlr-v139-schutt21a}
           & NequIP \cite{batzner_e3-equivariant_2022}
           & DetaNet \cite{zou_deep_2023}
           & EnviroDetaNet   \cite{xu_pretrained_2025}
           & MLANet                                                                                         \\
    \midrule
    Dipole Moment (D/\AA) & 0.0790                              & 0.0246 & 0.0102 & 0.0124  & 0.0087 & 0.0069 & 0.0188  \\
    Polarizability ($a^3_0$) & -                              & -  & 0.1536  & 0.1925   & 0.1374   & 0.0657   & 0.1745   \\
    \bottomrule
  \end{tabular}
  }
\end{table*}

To evaluate the force prediction capability of MLANet, we conducte tests on the MD17 dataset\cite{MD17}. First, we performe small-scale training by randomly selecting 950 samples for training, 50 for validation, and the remainder for testing. The force MAE results are presented in Table. S2 in the SI. MLANet demonstrates strong competitiveness, achieving leading MAE values across various small molecules, with the lowest error observed on benzene-containing systems.

\begin{figure}[htb]
\centering
\includegraphics[width=0.9\textwidth]{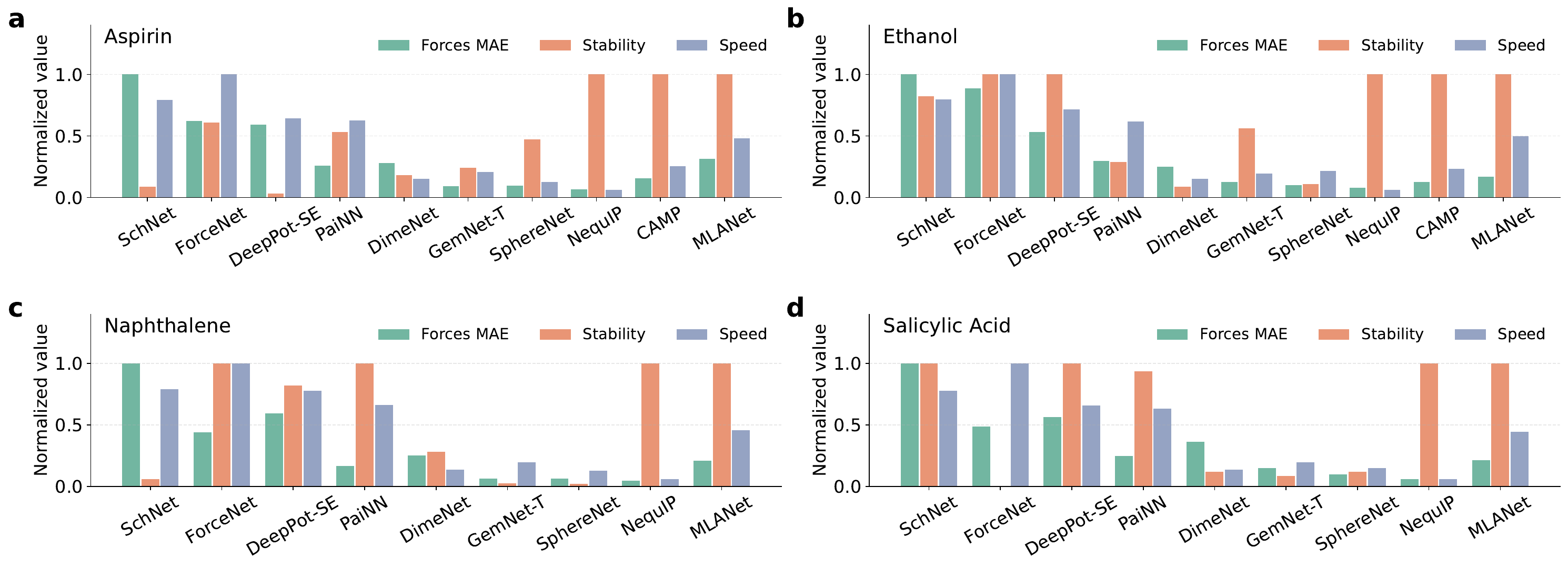}
\caption{\textbf{Performance comparison of force prediction accuracy (MAE), MD stability, and computational speed for small molecule systems.} Metrics are normalized to their maximum values for visualization. See Table. S3 in the SI for numerical data. Speed tests used a single NVIDIA V100 GPU.}
\label{fig6}
\end{figure}

Subsequently, we perform large-scale training using 9500 randomly selected samples for training, 500 for validation, and 10000 for testing. After model training, we select four molecules for molecular dynamics (MD) simulations to assess the stability and speed of MLANet. Fig. \ref{fig6} presents the results for aspirin, ethanol, malonaldehyde, and naphthalene (numerical values are provided in  Table. S3 in the SI). As shown, MLANet achieves relatively low force MAE, albeit slightly higher than models such as NequIP\cite{batzner_e3-equivariant_2022} and CAMP\cite{wen2025cartesian}. Simultaneously, it maintains stable simulations throughout the entire 300 ps duration. 
In terms of simulation speed, while not matching simpler scalar-based models, it significantly outperforms existing tensor-based models, even by an order of magnitude.

\subsection*{Two-dimensional system}
\begin{figure}[htb]
\centering
\includegraphics[width=0.9\textwidth]{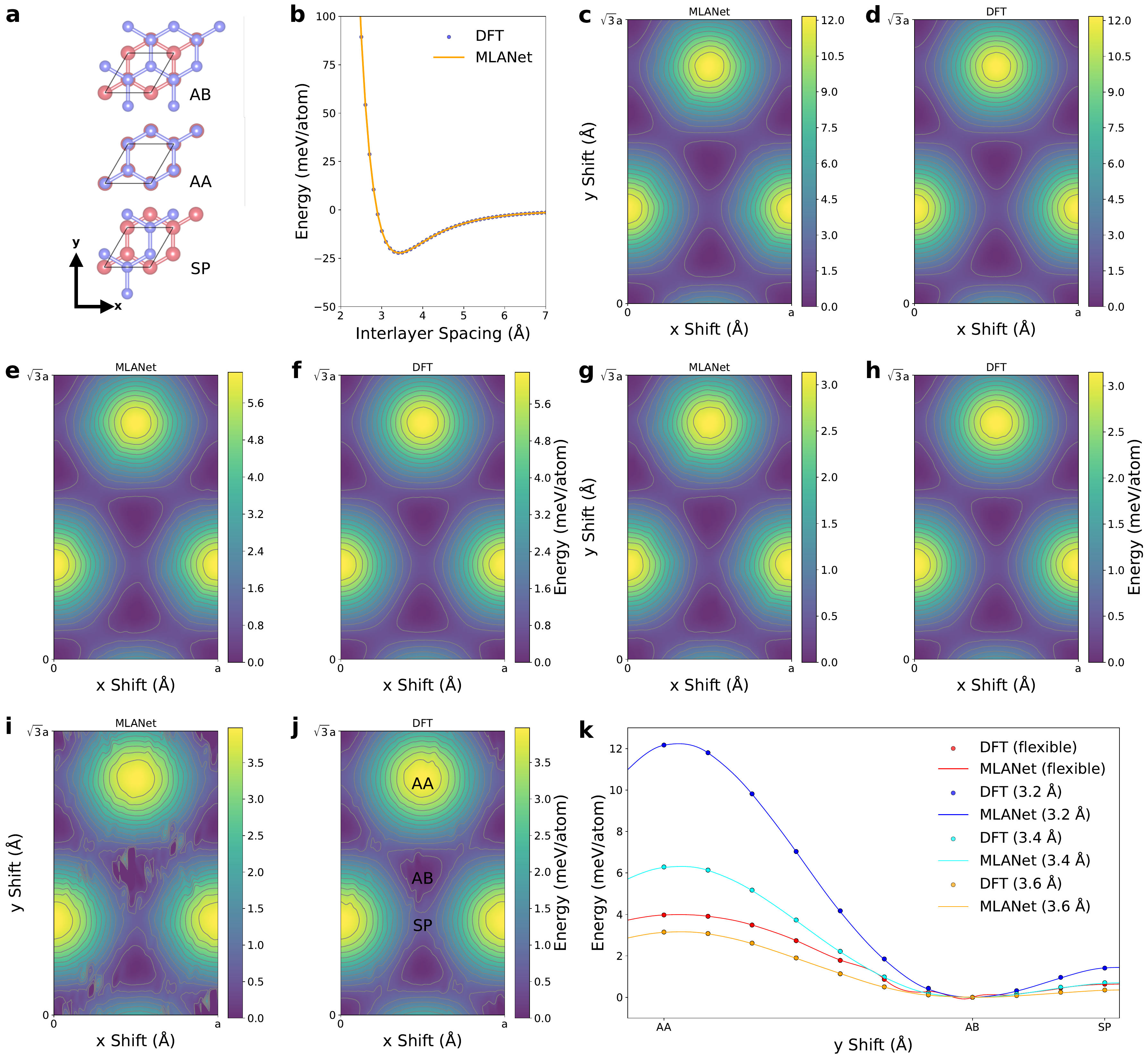}
\caption{\textbf{Comparison of MLANet predictions and DFT calculations.}
\textbf{a} AB, AA, and SP stacking configurations, whose positions are marked on the MLANet potential energy surface plot.
\textbf{b} Binding energy curves of AB-stacked bilayer graphene from DFT (blue) and MLANet (orange).
\textbf{c,\ d} Rigid-sliding potential energy surfaces of bilayer graphene from MLANet and DFT at a separation of 3.2 \AA.
\textbf{e,\ f} Rigid-sliding potential energy surfaces of bilayer graphene from MLANet and DFT at a separation of 3.4 \AA.
\textbf{g,\ h} Rigid-sliding potential energy surfaces of bilayer graphene from MLANet and DFT at a separation of 3.6 \AA.
\textbf{i,\ j} Flexible-sliding potential energy surfaces of bilayer graphene from MLANet and DFT at a separation of 3.2 \AA. The energy zero for both DFT and MLANet is set to the total energy of AB-stacked bilayer graphene.
\textbf{k} Energy scan curves along the AA-to-SP direction under four conditions from panel \textbf{j}, where the curves and data points represent predictions from MLANet and DFT, respectively.
}
\label{fig8}
\end{figure}

Two-dimensional (2D) materials play a significant role in material systems, exhibiting unique physical and chemical properties due to their partial periodicity. 
We evaluate MLANet on a bilayer graphene dataset\cite{ying_effect_2024}, focusing on its potential energy surfaces under different stacking configurations (Fig. \ref{fig8}a). Starting from the AB stacking configuration, the binding energy curve for bilayer graphene exfoliation predicted by MLANet aligns well with DFT results (Fig. \ref{fig8}b). Moreover, MLANet accurately predict the rigid-sliding potential energy surfaces (with fixed interlayer distances) at three different interlayer spacings (13.2 Å, 13.4 Å, and 13.6 Å) (Fig. \ref{fig8}c to \ref{fig8}h). For the flexible-sliding potential energy surface (where the interlayer distance adjusts with structural stability) at an interlayer spacing of 13.2 Å, MLANet successfully reproduce the overall energy landscape, despite some minor non-smooth energy points (Fig. \ref{fig8}i to \ref{fig8}j). 
In addition, the energy variation curves along the y-direction from AA to SP stacking for all four sliding conditions are also well reproduced by MLANet (Fig. \ref{fig8}k), showing good agreement with DFT predictions.

Furthermore, we present the RMSE of MLANet in predicting energies for sliding and exfoliation in bilayer graphene (Table. \ref{tab:dg}). MLANet achieve accurate results, ranking second only to the more structurally complex models NequIP and hNN.

\begin{table*}[h]
  \centering
  \caption{\textbf{RMSE values of binding energies and sliding energies for the potentials against reference DFT calculations in bilayer graphene.}}
  \label{tab:dg}
  \resizebox{\textwidth}{!}{
  \begin{tabular}{lllllllllll}
    \toprule
           & AIREBO \cite{10.1063/1.481208}
           & Reaxff \cite{reaxff}
           & GAP2020 \cite{10.1063/5.0005084}
           & LCBOP\cite{PhysRevB.68.024107}
           & DP \cite{WANG2018178}
           & NEP \cite{10.1063/5.0106617}
           & REBO-ILP \cite{rebo, ilp}
           & NequIP   \cite{batzner_e3-equivariant_2022}
           & hNN\cite{PhysRevB.100.195419}
           & MLANet                                                                                         \\
    \midrule
    Slinding Energy (meV/atom)& 1.1 & 0.7 & 2.5 & 0.9 & 0.6  & 0.7 & 1.0 & 0.4 & 0.4 & 0.5\\
    Binding Energy (meV/atom)& 17.7 & 16.0   & 10.3  & 7.4  & 3.5   & 2.6   & 1.6   & 1.2  & 0.8   & 1.3\\
    \bottomrule
  \end{tabular}
  }
\end{table*}

\subsection*{Formate decomposition}
To evaluate the performance of MLANet in surface catalytic systems, we test it on the formic acid decomposition dataset\cite{batzner_e3-equivariant_2022}. This dataset describes the decomposition of formic acid on a Cu surface ($\text{HCOO}^* \rightarrow \text{H}^* + \text{CO}_2$). The mean absolute errors (MAEs) for energy and forces are shown in the Table. \ref{tab:formate}.

\begin{table*}[h]
\centering
\caption{\textbf{Results on the Formate Decomposition datasets.} }
\resizebox{\textwidth}{!}{
\begin{tabular}{llllllll}
\toprule
\textbf{Dataset} & Metric & NequIP\cite{batzner_e3-equivariant_2022}  & MACE\cite{Batatia2022mace}& Mattersim\cite{yang2024mattersim}& Grace\cite{PhysRevX.14.021036}& AlphaNet\cite{yin2025alphanet}  & MLANet  \\ 
\midrule
\multirow{2}{*}{\textbf{Formate Decomposition}} & Force MAE (meV/\AA)  & 47.3 & 54.1  & 46.8& 53.9 & 42.5    & 44.9 \\ 
                                    & Energy MAE (meV/atom)  & 0.50 & 0.31  & 0.45 & 0.48 & 0.23   & 2.31 \\ 
\bottomrule
\end{tabular}
}
\label{tab:formate}
\end{table*}

On this dataset, MLANet achieve MAEs of 44.9 meV/\AA\  for forces and 2.31 meV/atom for energy. Although the energy error is relatively high, the low force error demonstrates MLANet's capability in simulating surface catalytic reactions with diverse interactions.
\subsection*{Water}
To evaluate the performance of MLANet in liquid systems, we test it on the bulk water dataset\cite{water}. The root mean square errors (RMSEs) for energy and forces are presented in the Table. \ref{tab:water}. 
MLANet achieve the best energy RMSE of 0.47 meV/atom, and its force RMSE of 60 meV/\AA\  is also competitive with other models. However, on this relatively small dataset consisting of only 1593 configurations, MLANet exhibits signs of overfitting. We believe that, as indicated by our earlier analysis, MLANet will demonstrate greater advantages on larger-scale datasets, where overfitting can be effectively mitigated, leading to improved robustness and higher accuracy.

\begin{table*}[h]
  \centering
  \caption{\textbf{Model performance on the water dataset.}
    RMSEs of energy and forces are in the units of meV/atom and meV/\AA, respectively.
  }
  \label{tab:water}
  \resizebox{\textwidth}{!}{
  \begin{tabular}{llllllllll}
    \toprule
           & BPNN \cite{water}
           & ACE \cite{PhysRevB.99.014104}
           & REANN \cite{PhysRevLett.127.156002}
           & DeePMD \cite{PhysRevLett.120.143001}
           & NequIP \cite{batzner_e3-equivariant_2022}
           & MACE \cite{Batatia2022mace}
           & CACE \cite{cheng2024cartesian}
           & CAMP  \cite{wen2025cartesian} 
           & MLANet                                                                                         \\
    \midrule
    Energy & 2.3                              & 1.7 & 0.8 & 2.1  & 0.94 & 0.63 & 0.59 & 0.59 & 0.47 \\
    Forces & 120                              & 99  & 53  & 92   & 45   & 36   & 47            & 34   & 60 \\
    \bottomrule
  \end{tabular}
  }
\end{table*}

\subsection*{Charged system}
On the charged system datasets\cite{ko_fourth-generation_2021}, compared to existing methods such as Qeq\cite{Rapp1991ChargeEF} and LES\cite{Kim2025Universal} that maintain charge conservation by incorporating long-range electrostatic interactions, MLANet demonstrates competitive performance in terms of force RMSE by directly embedding charge as a feature (Table. \ref{tab:charged}). In the C$_{10}$H$_2$/C$_{10}$H$_3^+$ and Ag$_3^{+/-}$ datasets, MLANet achieves lower RMSE than other direct charge-embedding approaches. Moreover, on the Na$_{8/9}$Cl$_8^+$ dataset, it surpasses models that include explicit long-range electrostatic interaction terms, attaining the lowest RMSE. Future integration of long-range electrostatic interactions into the direct force model is expected to yield even better performance.

\begin{table*}[h]
\centering
\caption{\textbf{Results on the different charge-state datasets}}
\resizebox{\textwidth}{!}{
\begin{tabular}{llllll}
\toprule
\textbf{Dataset} & \textbf{Metric} & 4G-HDNNP\cite{ko_fourth-generation_2021}  & Maruf’s NequIP\cite{Maruf2025EquivariantML}& ReaxNet\cite{gao_foundation_2025} & \textbf{MLANet } \\ 
\midrule
\textbf{C$_{10}$H$_2$/C$_{10}$H$_3^+$}
                                    & Force RMSE (eV/\AA)  & 0.078 & 0.071  & 0.023 & 0.074  \\ 
\textbf{Ag$_3^{+/-}$}
                                    & Force RMSE (eV/\AA)  & 0.033 & -  & 0.005 & 0.024  \\ 
\textbf{Na$_{8/9}$Cl$_8^+$}
                                    & Force RMSE (eV/\AA)  & 0.032 & 2.145  & 0.028 & 0.012  \\

\bottomrule
\end{tabular}
}
\label{tab:charged}
\end{table*}
\subsection*{Interface speed}

In addition to the MD speed test conducted on the NVIDIA V100 mentioned earlier, we also perform inference speed tests on an NVIDIA 4060 laptop with 8GB VRAM, representing a common user environment (Table. S4 in the SI). The results demonstrate that MLANet can simulate systems with over a thousand atoms even on 8GB VRAM, showcasing its excellent efficiency.

\section*{Methods}\label{sec11}
\subsection*{Datasets}
QM7\cite{blum, rupp}:The QM7 dataset is a subset of the GDB-13 database, which contains nearly one billion stable and synthetically accessible organic molecules. QM7 comprises molecules with up to 23 atoms, including seven heavy atoms (C, N, O, S), resulting in a total of 7,165 molecules. Following previous studies, we employ ten-fold cross-validation for testing.

Mptrj\cite{deng_2023_chgnet}:The MPtrj dataset comprises 1,580,395 structures, 1,580,395 energy values, 7,944,833 magnetic moments, 49,295,660 force entries, and 14,223,555 stress components. The structures and corresponding labels were parsed from all GGA/GGA+U static/relaxation trajectories in the Materials Project database (version 2022.9), employing a selection methodology designed to exclude incompatible calculations and duplicate structures. For model testing, we extracted all structures containing lithium (Li). The resulting training set consists of 219,033 structures, while the test set contains 1,449 structures.

SiO$_2$\cite{erhard_machine-learned_2022}:The SiO$_2$ dataset consists of 3074 structures, containing a total of 268,118 atoms. This dataset is generated using an iterative method, sometimes driven by empirical potentials.

GeSbTe\cite{zhou_device-scale_2023}:The GeSbTe dataset is calculated using the PBE functional. It covers a series of compositions along the pseudobinary line from $\text{GeTe}$ to $\text{Sb}_2\text{Te}_3$ and is created through a two-step iterative process. It contains 2692 structures, with a total of 341,132 atoms.

Phosphorus\cite{deringer_general-purpose_2020}:The phosphorus dataset contains structures generated by GAP-RSS, including liquid, crystalline, and isolated fragments. It consists of 4798 structures, with a total of 140,910 atoms. In this dataset, we used the average distance embedding between each atom and all other atoms to represent long-range interactions. 

QM9\cite{ramakrishnan2014quantum}:The QM9 dataset is constructed based on quantum chemical calculations, covering over 130,000 organic molecules composed of carbon, hydrogen, nitrogen, oxygen, and fluorine, with molecular weights not exceeding 900 Daltons. The dataset uses density functional theory (DFT) calculations to obtain quantum chemical properties such as geometric structures, energies, and charge distributions of the molecules in their ground states. During the construction process, researchers employed high-precision computational methods to ensure the accuracy and reliability of the data. We use 110,000 structures for training, 10,000 for validation, and the remainder for testing.

QM9S\cite{zou_deep_2023}:The QM9S dataset comprises approximately 130,000 organic molecules, encompassing a comprehensive set of quantum mechanical properties. These include molecular geometries, scalar properties (e.g., energy and partial charges), vector properties (such as dipole moments and atomic forces), second-order tensor properties (e.g., polarizability), and third-order tensor properties (e.g., hyperpolarizability). For consistency with prior work, we adopt the established data splitting scheme used in previous studies, dividing the dataset into training, validation, and test sets in a ratio of 90:5:5.

MD17\cite{MD17}:The MD17 dataset is a publicly available dataset for molecular dynamics research, containing energy and force information from molecular dynamics trajectories of eight organic molecules. It is designed to support the application of machine learning models in constructing accurate energy-conserving molecular force fields. We first use 950 structures for training, 50 for validation, and the remainder for testing on eight organic molecules. Then, on four of these organic molecules, we use 9,500 for training, 500 for validation, and 10,000 for testing.

Bilayer graphene\cite{ying_effect_2024}:The defective bilayer graphene dataset contains multiple structures with three stacking configurations: V0V0, V0V1, and V1V1. It includes DFT+MBD calculation results for various cases. The training, validation, and test sets consist of 3,988, 4,467, and 200 structures, respectively.

Formate decomposition\cite{batzner_e3-equivariant_2022}:The dataset for formate decomposition on a Cu surface contains 6,855 structures generated by the nudged elastic band (NEB) method and AIMD simulations. The training, validation, and test sets consist of 2,500, 250, and 4,105 structures, respectively.

Water\cite{water}:The water dataset contains 1,593 liquid water configurations, each consisting of 192 atoms, corresponding to 64 water molecules. These data are derived from density functional theory calculations at the revPBE0-D3 level. The dataset is split into training, validation, and test sets in a ratio of 90:5:5.

Charged system\cite{ko_fourth-generation_2021}:The C$_{10}$H$_2$/C$_{10}$H$_3^+$, Ag$_3^{+/-}$, and Na$_{8/9}$Cl$_8^+$ datasets contain 10,019, 11,013, and 5,000 structures, respectively. For each dataset, we use 90\% of the data for training and the remainder for validation.

\subsection*{Training details}

The model was implemented using PyTorch and trained on two NVIDIA GeForce RTX 4090 GPUs with 24GB memory each. The training objective combines weighted losses for energy ($E$), forces ($F$), predictions:

\begin{equation}
\mathcal{L}_{\text{total}} = \lambda_E\mathcal{L}_E + \lambda_F\mathcal{L}_F
\end{equation}

where $\lambda_E$ and $\lambda_F$ balance the relative importance of each term. The loss components utilize the L1 loss function for robust regression:

\begin{align*}
    \mathcal{L}_E &= \frac{1}{N} \sum_{i=1}^{N} \left| E_i - \hat{E}_i \right|, \\
    \mathcal{L}_F &= \frac{1}{3N} \sum_{i=1}^{N} \sum_{\alpha \in \{x,y,z\}} \left| F_{i,\alpha} - \hat{F}_{i,\alpha} \right|.
\end{align*}

The energy loss $\mathcal{L}_E$ computes the MAE between predicted and true energies across the batch, while the force loss $\mathcal{L}_F$ averages over all force components (3D vectors per atom).

The hyperparameters details in different datasets are shown in Table. S5 and S6 in the SI.

\section*{Discussion}\label{sec13} 

In this work, we presents MLANet, a graph neural network framework that integrates equivariant message passing with dynamic graph attention and multi-perspective pooling. Experimental results demonstrate that MLANet achieves a favorable balance among prediction accuracy, computational efficiency, and simulation stability, positioning it as a competitive new tool for large-scale atomic simulations.

MLANet demonstrates excellent predictive accuracy across diverse systems. It achieves competitive results on standard molecular benchmarks (e.g., QM7, MD17) and accurately models complex phenomena in materials, including phase transitions in SiO$_2$, intermetallic interactions in Ge-Sb-Te, and long-range forces in bilayer black phosphorus.
Crucially, MLANet combines this high accuracy with remarkable computational efficiency and scalability. Its performance scales favorably with data volume, and it offers practical guidance for balancing model complexity via maximum rotation order $l_{max}$ with data size and cost. In direct speed benchmarks, MLANet runs significantly faster—by up to an order of magnitude—than comparable models like NequIP, while maintaining simulation stability over long molecular dynamics trajectories (e.g., 300 ps for 4 molecules).

The success of MLANet stems from its architectural innovations. First, the geometry-aware dual-path dynamic attention mechanism explicitly decouples the importance assessment of messages from their feature transformation, significantly enhancing the model’s ability to discriminate subtle variations in local chemical environments. Second, the physics-informed multi-perspective pooling strategy—integrating additive, mean, and max operations—captures complementary feature information from extensive, intensive, and critical-local perspectives, leading to a more comprehensive and robust system representation. Together, these innovations enable the model to learn a highly accurate potential energy surface while maintaining markedly lower computational cost, a synergy that underpins its practical efficiency and generalization capability.

MLANet exhibits inherent computational efficiency as a direct-force model; however, its force prediction accuracy remains inferior to that of energy-conserving gradient-force models. This limitation restricts its applicability in tasks requiring high force-field smoothness and precise Hessian matrices, such as transition-state searches. Recent studies indicate that training direct-force models on large-scale datasets incorporating second-order derivative information (Hessian matrices) can substantially enhance the symmetry and accuracy of predicted forces, potentially enabling performance comparable to gradient-force models in reaction pathway exploration. Consequently, by integrating training data with higher-order physical observables, MLANet may achieve accuracy approaching that of gradient-force models while preserving computational efficiency, thereby evolving into a versatile tool for large-scale simulations that balances speed and precision.
In future work, we will enhance the MLANet framework through active learning for data generation, development of more efficient equivariant operators, and training on more comprehensive datasets that incorporate higher‑order physical information. These advancements are intended to establish MLANet as a robust computational tool for ultra-large-scale atomic simulations in critical domains such as catalysis and energy materials.

\section*{Data availability}
The datasets used in this paper are publicly available (see “Methods”).


\section*{Acknowledgements}
This work is supported by the National Natural Science Foundation of China (52574471, 52404423, 52334009), the Open Research Fund of Songshan Lake Materials Laboratory (2023SLABFK12), the Science and Technology Commission of Shanghai Municipality (23ZR1421600), and Research Project of State Key Laboratory of Advanced Special Steel (Shanghai)(SKLASS 2025-Z12). This work was 
also supported by Shanghai Technical Service Center of Science and 
Engineering Computing, Shanghai University.
\section*{Author contribution}
Shuyu Bi: Writing–original draft, Writing–review \& editing, Software, Methodology, Investigation, Formal analysis, Data curation. Zhede Zhao: Writing–review \& editing. Qiangchao Sun: Supervision. Tao Hu: Writing–review \& editing, Investigation, Supervision, Resources. Xionggang Lu: Resources. Hongwei Cheng: Supervision, Resources, Project administration, Investigation, Funding acquisition.
\section*{Competing interests}
The authors declare no competing interests.



\end{document}


\title{\Large Supplementary Information: \\ Universal and efficient graph neural networks with dynamic attention for machine learning interatomic potentials\\}

\author{Shuyu Bi$^1$}
\author{Zhede Zhao$^1$}
\author{Qiangchao Sun$^1$}
\author{Tao Hu$^1$}
\email{taohu@shu.edu.cn}
\author{Xionggang Lu$^1$}
\author{Hongwei Cheng$^1$}
\email{hwcheng@shu.edu.cn}
\affiliation{$^1$School of Materials Science and Engineering \& State Key Laboratory of Advanced
Refractories \& State Key Laboratory of Advanced Special Steels, Shanghai University, Shanghai 200444, P. R. China. 
}

\maketitle

\section{Results on Mptrj-Li}
\begin{figure}[H]
    \centering
    \includegraphics[width=0.9\textwidth]{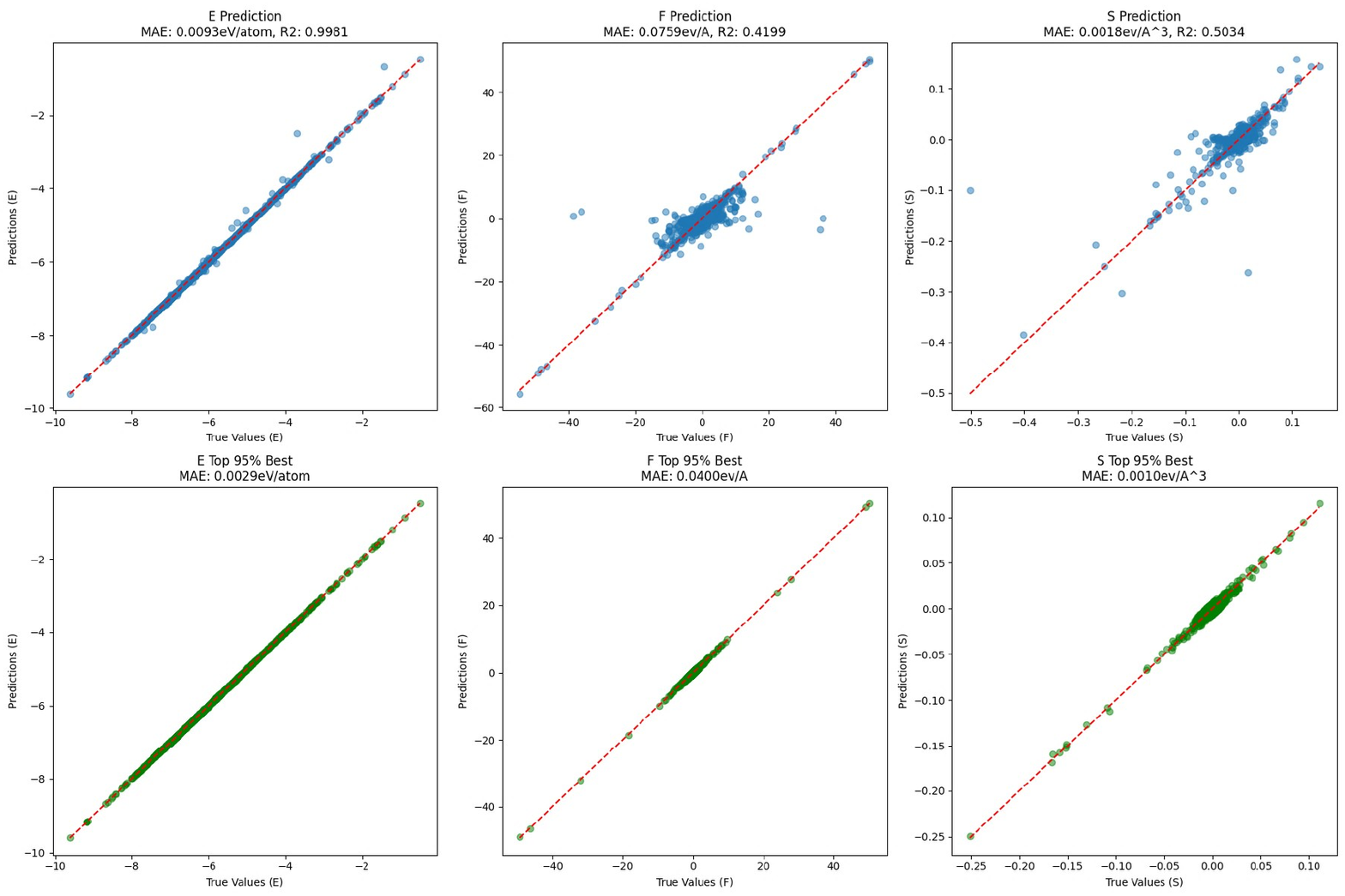}
    \caption{\textbf{The prediction error distributions and fitted scatter plots for energy, interatomic forces, and stress on the Li-containing subset of MPtrj.} The green region highlights the 95\% of structures with the lowest prediction errors, emphasizing the model's performance within the majority of the data range.
    }
    \vspace{150pt}  
    \label{fig:s1}
\end{figure}

\section{Results on SiO$_2$}
\begin{figure}[H]
    \centering
    \includegraphics[width=0.9\textwidth]{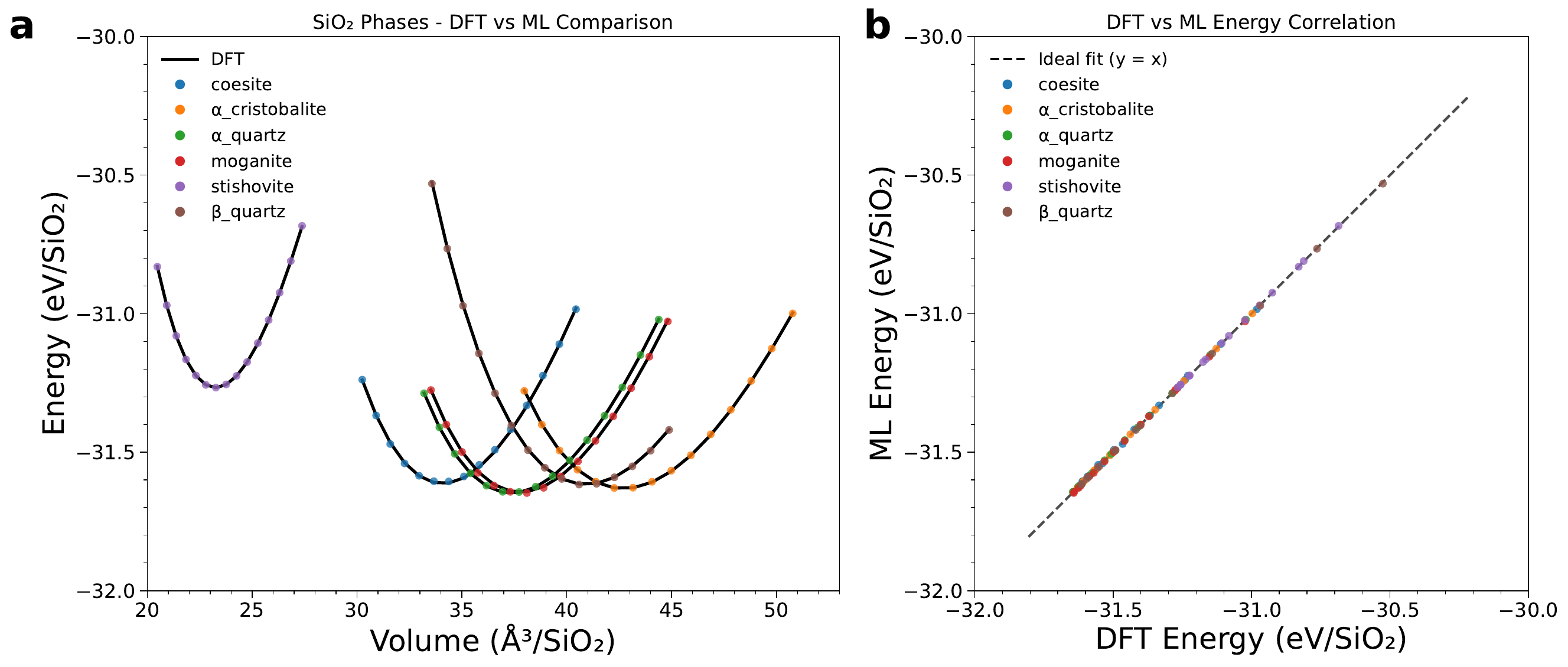}
    \caption{\textbf{Energy–volume curves and error scatter plots for different silica polymorphs.} \textbf{a, b} The colored data points correspond to calculation results using the MLANet model developed in this work. MLANet shows good agreement with the SCAN-DFT data (solid black line).}
    \label{fig4}
    \vspace{250pt}  
\end{figure}

\section{Results on Ge–Sb–Te}
\begin{figure}[H]
    \centering
    \includegraphics[width=0.9\textwidth]{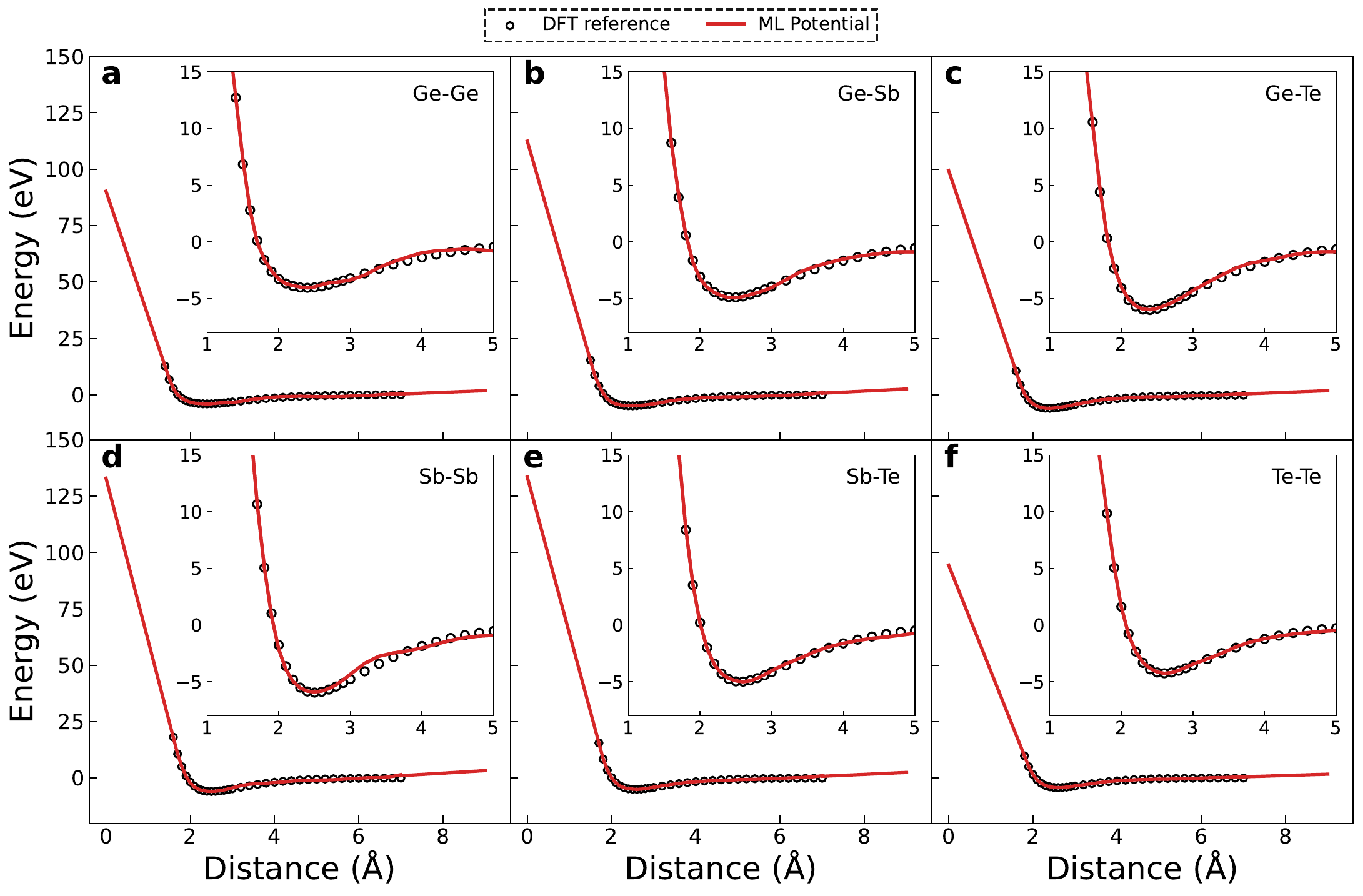}
    \caption{\textbf{Potential energy scans for pairwise atomic interactions in the Ge–Sb–Te system.} Results are shown for isolated dimers across all possible atomic pair types: \textbf{a} Ge–Ge, \textbf{b} Ge–Sb, \textbf{c} Ge–Te, \textbf{d} Sb–Sb, S\textbf{e} b–Te, and \textbf{f} Te–Te.
    }
    \vspace{150pt}  
    \label{fig:s1}
\end{figure}

\section{Results on QM9}
\begin{table}[H]{
    \centering
    \caption{\textbf{MAE on QM9 property prediction. }}
    \label{tab:qm9}
    \resizebox{\textwidth}{!}{
    \begin{tabular}{lllllllll}
    \toprule
    & $\mu$ & $\alpha$ & homo & lumo & gap & $R^2$ & ZPVE & $C_v$  \\
    & (D) & ($a_0^3$) & (meV) & (meV) & (meV) & ($a_0^2$) & (meV) & ($\tfrac{\text{cal}}{\text{mol}\cdot \text{K}}$)\\
    \midrule
    SchNet\cite{schutt_schnet_2018}         & 0.033 & 0.235 & 41.0 & 34.0 & 63.0 & 0.070 & 1.70  & 0.033 \\
    EGNN  \cite{satorras2022enequivariantgraphneural}        & 0.029 & 0.071 & 29.0 & 25.0 & 48.0 & 0.106 & 1.55  & 0.031 \\
    DimeNet++ \cite{gasteiger_dimenetpp_2020}    & 0.030 & 0.044 & 24.6 & 19.5 & 32.6 & 0.330 & 1.21   & 0.023 \\
    PaiNN  \cite{pmlr-v139-schutt21a}       & 0.016 & 0.045 & 27.6 & 20.4 & 45.7 & 0.070 & 1.28   & 0.024 \\
    SphereNet \cite{Liu2021SphericalMP}    & 0.025 & 0.045 & 22.8 & 18.9 & 31.1 & 0.270 & 1.12   & 0.022  \\
    MEGNet \cite{battaglia2018relationalinductivebiasesdeep}    & 0.050 & 0.081 & 43.0 & 44.0 & 66.0 & 0.302 & 1.43   & 0.029  \\
    enn-s2s  \cite{pmlr-v70-gilmer17a}   & 0.030 & 0.092 & 43.0 & 37.0 & 69.0 & 0.180 & 1.50   & 0.040  \\
    ALIGNN \cite{choudhary_atomistic_2021}    & 0.0146 & 0.0561 & 21.4 & 19.5 & 38.1 & 0.106 & 3.10   & -  \\
    TorchMD-Net \cite{tholke2021equivariant}  & 0.011 & {0.059} & 20.3 & 17.5 & 36.1 & 0.033 & 1.84   & 0.026  \\
    MLANet    & 0.024 & 0.087 & 46.7 & 36.7 & 72.7 & 0.277 & 1.53   & 0.034  \\
    \bottomrule
    \end{tabular}
    }}
    \vspace{200pt}
    \end{table}

\section{Results on MD17}

\begin{table}[H]
    \centering
    \caption{\textbf{MAE of force prediction on MD17 dataset (kcal/mol/Å).} }
    \label{tab:md17}
    \resizebox{\textwidth}{!}{
    \begin{tabular}{lllllllll}
    \toprule
    & Aspirin & Benzene & Ethanol & Malonaldehyde & Naphthalene & Salicylic Acid & Toluene & Uracil \\
    \midrule
    SphereNet  \cite{Liu2021SphericalMP}                       & 0.430 & 0.178 & 0.208 & 0.340 & 0.178 & 0.360 & 0.155 & 0.267 \\
    SchNet \cite{schutt_schnet_2018}                           & 1.350 & 0.310 & 0.390 & 0.660 & 0.580 & 0.850 & 0.570 & 0.560 \\
    DimeNet \cite{gasteiger_dimenet_2020}                          & 0.499 & 0.187 & 0.230 & 0.383 & 0.215 & 0.374 & 0.216 & 0.301 \\
    PaiNN  \cite{pmlr-v139-schutt21a}                           & 0.338 & --    & 0.224 & 0.319 & 0.077 & 0.195 & 0.094 & 0.139 \\
    PhysNet \cite{Unke2019PhysNetAN}                            & 0.605 & --    & 0.160 & 0.319 & 0.310 & 0.337 & 0.191 & 0.218 \\
    TorchMD-Net \cite{tholke2021equivariant}                      & 0.245 & 0.219 & 0.107 & 0.167 & 0.059 & 0.128 & 0.064 & 0.089 \\
    MLANet    & 0.425 & 0.134 & 0.284 & 0.295 & 0.162 & 0.281 & 0.204 & 0.329 \\
    \bottomrule
    
    \end{tabular}
    }
    \vspace{400pt}
    \end{table}

\section{Results on 4 moleculars of MD17}

\begin{table}[H]
    \centering
    \small
    \caption{
        \textbf{Results of Aspirin, Ethanol, Naphthalene and Salicylic Acid trained on 9500 structures.}
        Force MAE in meV/\AA, stability measured in picoseconds, speed measured in frames per second.
        Results from all models (DeepPot-SE\cite{82e463d22b5a4d52ac8f62989503daf0} , SchNet\cite{schutt_schnet_2018}, DimeNet\cite{gasteiger_dimenet_2020}  , PaiNN\cite{pmlr-v139-schutt21a}   , SphereNet\cite{Liu2021SphericalMP} , ForceNet\cite{hu2021forcenetgraphneuralnetwork}, GemNet-T\cite{gasteiger_gemnet_2021} , and NequIP\cite{batzner_e3-equivariant_2022} ) except for CAMP\cite{wen2025cartesian}  and MLANet, are obtained from benchmark study in ref. \cite{fu_forces_2023}.
    }
    \resizebox{\textwidth}{!}{
    \begin{tabular}{lllllllllllllll}
        \hline
        Molecule & Metric    & DeepPot-SE  & SchNet        & DimeNet     & PaiNN      & SphereNet     & ForceNet      & GemNet-T     & NequIP      & CAMP     & MLANet      \\
        \hline
        Aspirin  & Forces    & 21.0        & 35.6          & 10.0        & 9.2           & 3.4          & 22.1          & 3.3          & 2.3         & 5.5         & 11.1         \\
                 & Stability & 9$_{(15)}$  & 26$_{(23)}$   & 54$_{(12)}$ & 159$_{(121)}$ & 141$_{(54)}$ & 182$_{(144)}$ & 72$_{(50)}$  & 300$_{(0)}$ & 300$_{(0)}$ & 300$_{(0)}$ \\
                 & Speed     & 88.0        & 108.9         & 20.6        & 85.8          & 17.5         & 137.3         & 28.2         & 8.4         & 35.0        & 65.8        \\
        Ethanol  & Force     & 8.9         & 16.8          & 4.2         & 5.0           & 1.7          & 14.9          & 2.1          & 1.3         & 2.1         & 2.7         \\
                 & Stability & 300$_{(0)}$ & 247$_{(106)}$ & 26$_{(10)}$ & 86$_{(109)}$  & 33$_{(16)}$  & 300$_{(0)}$   & 169$_{(98)}$ & 300$_{(0)}$ & 300$_{(0)}$ & 300$_{(0)}$ \\
                 & Speed     & 101.0       & 112.6         & 21.4        & 87.3          & 30.5         & 141.1         & 27.1         & 8.9         & 32.8        & 70.0        \\
          Naphthalene   & Force     & 13.4         & 22.5          & 5.7         & 3.8           & 1.5          & 9.9          & 1.5          & 1.1         & -         & 4.7         \\
                        & Stability & 246$_{(109)}$ & 18$_{(2)}$ & 85$_{(68)}$ & 300$_{(0)}$  & 6$_{(3)}$  & 300$_{(0)}$   & 8$_{(2)}$ & 300$_{(0)}$ & - & 300$_{(0)}$ \\
                        & Speed     & 109.3       & 110.9         & 19.1       & 92.8         & 18.3         & 140.2         & 27.7         & 8.2         & -        & 64.1       \\
        Salicylic Acid  & Force     & 14.9         & 26.3          & 9.6         & 6.5           & 2.6          & 12.8          & 4.0          & 1.6         & -         & 5.6         \\
                        & Stability & 300$_{(0)}$ & 300$_{(0)}$ & 73$_{(82)}$ & 281$_{(37)}$  & 36$_{(16)}$  & 1$_{(0)}$   & 26$_{(24)}$ & 300$_{(0)}$ & - & 300$_{(0)}$ \\
                        & Speed     & 94.6        & 111.7        & 19.4        & 90.5          & 21.4         & 143.2         & 28.5         & 8.4         & -        & 63.9        \\
        \hline
    \end{tabular}
    }
    \vspace{250pt}
\end{table}

\section{Interface speed}
\begin{table*}[h]
    \caption{Inference Speed Test of the Model (Unit: ms). The test structures used were diamond and its supercells, with "OOM" indicating out of memory.}
    \begin{tabular}{lcccc}
        \hline
        model & 27 atoms  & 64 atoms  & 512 atoms & 1000 atoms  \\
        \hline
        
        MACE\cite{Batatia2022mace} & 328.4699 & 931.7864 & OOM & OOM \\
        ORB-V3\cite{rhodes2025orbv3atomisticsimulationscale} & 185.6557 & 436.2073 & OOM & OOM \\
        SevenNet-0\cite{park_scalable_2024} & 192.0174 &  197.8365  & OOM & OOM \\
        SevenNet-i3i5\cite{park_scalable_2024} & 591.3817 &  OOM  & OOM & OOM \\
        CHGNet\cite{deng_2023_chgnet} & OOM & OOM & OOM & OOM \\
        MLANet & 21.5072 & 25.5946 & 357.3566 & 1437.9338 \\

        \hline
        \end{tabular}
        \vspace{250pt}
    \end{table*}
\section{Training details}
\begin{table}[H]
    \centering
    \small
    \caption{\textbf{Training hyperparameters used in different datasets.}}
    \label{tab:trainparams}
    \resizebox{\textwidth}{!}{
    \begin{tabular}{llllll}
        \toprule
        Dataset & Learning rate & Optimizer & Batch size & Energy\&force ratio &scheduler \\
        \midrule
        
        QM7 & 4e-4 & AdamW & 128 & 1:0 & CosineAnnealingLR\\
        QM9/QM9S & 4e-4 & AdamW & 128 & 1:0 & CosineAnnealingLR\\
        MD17 & 4e-4 & AdamW & 128 & 0:1000 & CosineAnnealingLR\\
        MPtrj-Li & 2e-3 & AdamW & 200 & 1:1000 & CosineAnnealingLR\\
        SiO$_2$ & 4e-4 & AdamW & 32 & 1:1000 & CosineAnnealingLR\\
        GeSbTe & 4e-4 & AdamW & 32 & 1:1000 & CosineAnnealingLR\\
        Phosphorus & 4e-4 & AdamW & 32 & 1:1000 & CosineAnnealingLR\\
        Bilayer graphene & 4e-4 & AdamW & 32 & 1:1000 & CosineAnnealingLR\\
        Formate decomposition & 4e-4 & AdamW & 32 & 1:1000 & CosineAnnealingLR\\
        Water & 4e-4 & AdamW & 16 & 1:1000 & CosineAnnealingLR\\
        Charged system & 4e-4 & AdamW & 128 & 1:1000 & CosineAnnealingLR\\
        \bottomrule
    \end{tabular}}
    \vspace{300pt}
\end{table}

\begin{table}[H]
    \centering
    \small
    \caption{\textbf{Architectural hyperparameters used in different datasets.} l$_E$, l$_F$ and l$_{MLP}$ represent Energy message passing layers, Force message passing layers and Multilayer Perceptron layers. }
    \label{tab:arparams}
    \resizebox{\textwidth}{!}{
    \begin{tabular}{llllllll}
        \toprule
        Dataset & L$_{max}$ &  l$_E$ &  l$_F$ & l$_{MLP}$  & Cutoff $r_cut$ (\AA)& Activation & Hidden irreps\\
        \midrule
        
        QM7 & 3 & 1 & 0 & 1 & 5.0 & SILU & 128x0e+64x1o\\
        QM9/QM9S & 3 & 1 & 0 & 1 & 6.0 & SILU & 128x0e+64x1o\\
        MD17 & 3 & 1 & 4 & 1 & 6.0 & SILU & 128x0e+64x1o+32x2e+32x3o\\
        MPtrj-Li & 3 & 1 & 2 & 4 & dynamic & SILU & 128x0e+128x1o+128x2e+32x3o\\
        SiO$_2$ & 3 & 1 & 0 & 4 & 5.0 & SILU & 128x0e+64x1o+32x2e+32x3o\\
        GeSbTe & 3 & 1 & 0 & 4 & 5.0 & SILU & 128x0e+64x1o+32x2e+32x3o\\
        Phosphorus & 3 & 1 & 0 & 4 & 5.0 & SILU & 128x0e+64x1o+32x2e+32x3o\\
        Bilayer graphene & 3 & 1 & 0 & 4 & 5.0 & SILU & 128x0e+64x1o+32x2e+32x3o\\
        Formate decomposition & 3 & 1 & 2 & 4 & 5.0 & SILU & 128x0e+64x1o+32x2e+32x3o\\
        Water & 3 & 1 & 2 & 4 & 5.0 & SILU & 128x0e+64x1o+32x2e+32x3o\\
        C$_{10}$H$_2$/C$_{10}$H$_3^+$ & 3 & 1 & 2 & 4 & 4.23 & SILU & 128x0e+64x1o+32x2e+32x3o\\
        Ag$_3^{+/-}$ & 3 & 1 & 2 & 4 & 5.29 & SILU & 128x0e+64x1o+32x2e+32x3o\\
        Na$_{8/9}$Cl$_8^+$ & 3 & 1 & 2 & 4 & 5.29 & SILU & 128x0e+64x1o+32x2e+32x3o\\
        \bottomrule
    \end{tabular}}
    \vspace{350pt}
\end{table}

\def\bibsection{\section*{\refname}} 
%